\documentclass[letterpaper, 10 pt, conference]{ieeeconf}  %

\IEEEoverridecommandlockouts                              %
\usepackage{booktabs}
\usepackage{hyperref}
\usepackage{graphicx} 
\usepackage{amsmath} 
\usepackage{amssymb}
\newtheorem{remark}{Remark}
\usepackage{comment}
\usepackage{multirow}
\usepackage{color}
\usepackage{url}
\usepackage{subcaption}
\usepackage{float}

\newcommand{\tr}{\mathrm{tr}}
\newcommand{\te}{\mathrm{te}}

\newcommand{\norm}[1]{\left\lVert#1\right\rVert}

\newcommand{\seq}[1]{{\mathbf{#1}}}

\newcommand{\D}{D}
\newcommand{\E}{\mathbb{E}} 

\newcommand{\redpar}{\phi}
\newcommand{\projpar}{\gamma}
\newcommand{\encpar}{\psi}

\newcommand{\diag}{\mathrm{diag}}
\newcommand{\loss}{\mathcal{L}}

\newcommand{\M}{F}

\title{\LARGE \bf Variational meta-learning inference for low dimensional neural system identification
}

\author{Matteo Rufolo$^{1}$, Dario Piga$^{1}$, Marco Forgione$^{1}$
\thanks{Corresponding author: {\tt matteo.rufolo@supsi.ch}.}
\thanks{$^{1}$SUPSI-DTI-IDSIA, Dalle Molle Institute for Artificial Intelligence, Lugano, Switzerland.}}

\begin{document}

\maketitle
\thispagestyle{empty}
\pagestyle{empty}

\begin{abstract}
Deep learning has proven highly effective for nonlinear system identification, but heavily parameterized neural networks are prone to overfitting in low-data regimes and lack reliable uncertainty quantification. The recently developed manifold meta-learning framework addresses the data efficiency problem by restricting the model parameters to a meta-learned low-dimensional manifold. However, that method is purely deterministic. We propose a fully probabilistic extension of the manifold meta-learning framework, based on amortized Variational Inference, where a generative prior over the low-dimensional parameter manifold is learned. During task-specific adaptation, we combine Maximum A Posteriori estimation with the Laplace approximation to yield a mathematically grounded posterior approximation. Evaluated on a static regression task and the Bouc--Wen dynamical system benchmark, the proposed approach achieves predictive accuracy comparable to its deterministic counterpart while successfully providing calibrated uncertainty bounds in severely low-data regimes.
\end{abstract}

\section{Introduction}
In recent years, system identification has increasingly adopted deep learning techniques, leveraging the high expressivity of black-box models such as neural networks to capture complex, nonlinear dynamics when the underlying physics are unknown~\cite{pillonetto2025deep}. 
However, a primary challenge in deploying these heavily parameterized models is their reliance on large datasets to ensure reliability across diverse operating conditions. When data is scarce, training these architectures becomes prone to overfitting and poor generalization. This contrasts with physics-based models, which are typically data-efficient and capable of generalizing beyond the training regime, though deriving such first-principles models is often unfeasible for complex systems.

Meta-learning techniques have recently gained traction within the system identification
and control communities to facilitate rapid adaptation across related tasks, spanning
distributionally robust task weighting~\cite{rufolo2025distributionally}, probabilistic
transformer system identification~\cite{rufolo2025enhanced}, transformer-based filtering~\cite{du2023can}
and controller-tuning transfer~\cite{lakshminarayanan2025inverse}. A key architectural
question shared across this line of work is how to summarize a variable-length dataset
into a fixed-size task representation for conditioning: permutation-invariant set encoders
such as Deep Sets~\cite{zaheer2017deep} and the Set Transformer~\cite{lee2019set} enforce
exchangeability when the ordering of observations carries no information, while sequence
encoders such as (bidirectional) recurrent networks are preferred when temporal ordering
is itself informative. This amortized-encoding paradigm also underlies the neural process
family~\cite{garnelo2018conditional}, which learns to map a context set directly to a
predictive distribution rather than solving an explicit inner optimization per task.

In particular, this paper extends a recently proposed manifold meta-learning
framework~\cite{forgione2025manifold}, which leverages an amortized learning strategy,
instantiated as a permutation-invariant or sequence encoder depending on whether the
task data is order-invariant or sequential, as discussed above, to automatically
extract simplified, efficient model architectures. The approach relies on an
over-parameterized \emph{base architecture} capable of capturing the dynamics of all
systems within a meta-dataset. Since the physical factors varying across systems
correspond to far fewer degrees of freedom than the total parameter count, it is natural
to seek a low-dimensional manifold within the base architecture's parameter space.
Restricting the model parameters to this meta-learned manifold yields a
reduced-complexity representation that achieves performance comparable to the full
model in data-rich scenarios, while substantially outperforming it when training data
is scarce. Unlike neural processes, which predict directly from the amortized
representation, the manifold framework retains an explicit low-dimensional parameter
$\phi$ and lifting function, using the encoder's output only to initialize, rather
than replace, a subsequent optimization-based refinement (Section~\ref{sec:MAP}).

The primary contribution of this work is the extension of~\cite{forgione2025manifold} into a probabilistic setting, drawing on recent advances in probabilistic machine learning~\cite{rufolo2025enhanced,garnelo2018conditional}. Specifically, we adopt a variational learning approach inspired by the Variational Autoencoder (VAE) framework~\cite{kingma2013auto}, in which an inference network processes each training dataset and outputs an approximate Gaussian posterior over the model's reduced-complexity parameters in the low-dimensional manifold. Despite the Gaussian restriction on the latent space, the highly nonlinear transformation from the manifold to the full-complexity model output can induce arbitrarily complex, non-Gaussian distributions over the predicted outputs. This probabilistic formulation enables rigorous uncertainty quantification and enhances model interpretability without sacrificing predictive accuracy.


The remainder of the paper is organized as follows. Section~\ref{sec:manifold} reviews the deterministic manifold meta-learning framework. Section~\ref{sec:vae_extension} details the mathematical formulation, training, and testing procedures for the proposed probabilistic extension. Finally, Section~\ref{sec:results} presents two numerical examples: a static regression and a dynamical system identification problem, demonstrating the framework's capabilities in uncertainty quantification and interpretability while maintaining similar performance to its deterministic counterpart.


\section{Manifold Meta-Learning for Neural System Identification}
\label{sec:manifold}
\subsection{Data Distribution}
In line with the meta-learning setting for system identification, we assume access to a meta-dataset $\mathcal{D} = \{\D^{(i)}\}_{i\in\mathbb{N}}$, where each dataset $\D^{(i)} = (\mathbf{u}^{(i)}, \mathbf{y}^{(i)})$ consists of input-output sequences of length $N$ generated by a specific dynamical system $S^{(i)}$.

We assume the underlying data-generating mechanism across datasets is governed by a low-dimensional latent variable $\mathbf{z} \in \mathbb{R}^{n_z}$. This variable  corresponds to the abstract factors of variation across systems (e.g., varying payload masses in a robotic arm or varying friction coefficients). The inputs $\mathbf{u}^{(i)}$ are independently sampled from a distribution $p(\mathbf{u})$, while the outputs depend on both the inputs and the system-specific latent variable via $p(\mathbf{y} \mid \mathbf{u}, \mathbf{z})$. Overall, the collection of datasets can be formalized as sampling from:
\begin{equation}
\label{eq:data_distrib}
p(\D) = p(\mathbf{u})\int p(\mathbf{z})p(\mathbf{y} \mid \mathbf{u}, \mathbf{z}) d\mathbf{z}.
\end{equation}

\subsection{Manifold Meta-Learning and Amortized Optimization}
The goal of~\cite{forgione2025manifold} is to learn a model structure of the input-output dependency 
$\mathbf{u} \mapsto \mathbf{y}$ across datasets  governed by a low-dimensional parameter vector $\phi$, consistent with the low-dimensionality hypothesis in~\eqref{eq:data_distrib}. Concretely, this is 
achieved by restricting the weights $\theta$ of an overparameterized neural network, referred to as~\emph{base architecture}:
\begin{equation}
\hat {\mathbf{y}} = F(\mathbf{u}, \theta), \qquad \theta \in \mathbb{R}^{n_\theta}
\end{equation}
to a low-dimensional manifold via a learned \emph{lifting function}
$P : \mathbb{R}^{n_\phi} \to \mathbb{R}^{n_\theta}$, with $n_\phi \ll n_\theta$. This yields the reduced-complexity model structure:
\begin{equation}
    \hat{\mathbf{y}} = F\!\left(\mathbf{u},\, P(\phi)\right)
    \label{eq:manifold_model}, \qquad \phi \in \mathbb{R}^{n_\phi}.
\end{equation}
%




In~\cite{forgione2025manifold}, the lifting function is parameterized: $P = P_\gamma$, where $\gamma \in \mathbb{R}^{n_\gamma}$ are meta-parameters. The latter are meta-learned across datasets to ensure that each input-output dependency in $\mathcal{D}$ can be accurately approximated by $F(\mathbf{u}, P_\gamma(\phi))$, for some dataset-dependent $\phi$ and for a shared $\gamma$.
This is equivalent to meta-learning a low-dimensional manifold $\mathcal{M} = \{\theta \in \mathbb{R}^{n_{\theta}} : \theta = P_\gamma(\phi), \phi \in \mathbb{R}^{n_{\phi}}\}$ in base architecture's parameter space that preserves the base architecture's approximation capabilities within the meta-dataset. By splitting each dataset $D$ into a training  portion $D_\tr = {(\mathbf{u}_{\tr}, \mathbf{y}_{\tr})}$ of length $N_\tr$, and a test portion $D_\te = {(\mathbf{u}_{\te}, \mathbf{y}_{\te})}$ of length $N_\te$ (with $N=N_\tr + N_\te$), the objective is formalized as a bi-level optimization problem:
\begin{subequations}
\label{eq:bilevel}
\begin{align}
\hat \gamma &= \arg \min_{\gamma} \E_{p(\D)} \big [ \loss({\seq y}_\te, F\left(\seq{u}_\te ; \, P_{\gamma} ( \hat \redpar)\right) \big ] \label{eq:outer}\\
\hat \redpar &=\arg \min_\phi \mathcal{L}(\seq y_\tr, \M(\seq u_\tr ; \, P_\projpar(\redpar))), \label{eq:inner}
\end{align}
\end{subequations}
where $\mathcal{L}$ is the mean squared error loss. In essence, \eqref{eq:bilevel} explicitly seeks the manifold that maximizes the average test performance, when fitting of reduced-complexity models is done by optimizing $\mathcal{L}$ on each training split.


Because exact bi-level optimization is computationally prohibitive, the framework employs an amortized optimization procedure \cite{amos2023tutorial}. The inner optimization \eqref{eq:inner} is replaced by an encoder network $E_\psi$ that directly predicts the low-dimensional parameter $\phi$ from the training data. This allows for end-to-end meta-learning according to the criterion:
\begin{equation}
\label{e1:amortized_train}
    \hat \projpar, \hat \encpar = \arg \min_{\projpar, \encpar} \E_{p(\D)} \big [ \loss({\seq y}_\te, F(
  \seq{u}_\te ; \,
  P_{\projpar} ( E_{\psi}(\seq u_\tr , \seq y_\tr ))) \big ],
\end{equation}
with the expectation approximated by the sample average over datasets in $\mathcal{D}$. 

At test time, the encoder $E_{\hat \psi}$ provides a fast, zero-shot estimation of $\phi$. However, because $E_{\hat \psi}$ is trained to approximate the solution of a nonconvex inner optimization \eqref{eq:inner} via a single amortized forward pass, its output is treated throughout this paper as an initializer rather than a final estimate: at meta-test time (Section \ref{sec:MAP}) it is always followed by gradient-based MAP refinement, so the encoder's role is to place the optimizer in the right basin of attraction rather than to be trusted in isolation. via gradient descent.

\textbf{Limitations of the Deterministic Framework:}
While this manifold meta-learning strategy successfully reduces model complexity, it is fundamentally \textit{deterministic}. The learned encoder $E_{\hat\psi}$ maps a dataset into a single point estimate $\phi$ on the manifold, failing to quantify the epistemic uncertainty inherent in limited-data scenarios. In practical system identification, especially for safety-critical applications, knowing the uncertainty in both the parameter estimates ($\phi$) and the resulting output predictions ($\hat{\mathbf{y}}$) is paramount. This fundamental limitation motivates 
a probabilistic reformulation of the manifold meta-learning discussed so far. 

\section{Variational Extension: Formulation}
\label{sec:vae_extension}

\subsection{Learning the probability of the latent variable}
To explicitly account for the uncertainty lacking in the deterministic framework, we transition to a generative probabilistic model. 
To this aim, we introduce \emph{learnable}  prior $p_\omega(\phi)$ and likelihood $p_\gamma(\mathbf{y}^{(i)} \mid \mathbf{u}^{(i)}, \phi)$ with meta-parameters $\omega$ and $\gamma$, respectively.
The likelihood is chosen as:
\begin{equation}
     p_\gamma(\mathbf{y}^{(i)} \mid \mathbf{u}^{(i)}, \phi) = \mathcal{N}(F(\mathbf{u}^{(i)}, P_\gamma(\phi)), \sigma_e^2),
\end{equation}
i.e., a Gaussian whose mean has the same structure of the point prediction in the deterministic manifold framework, and characterized by a noise variance $\sigma_e^2$. \footnote{In this section, the noise variance $\sigma_e^2$ is assumed to be known to simplify notation and derivations. In the numerical experiments, $\sigma_e^2$ is learned together with the other likelihood parameters $\gamma$.}
The prior $p_\omega(\phi)$ is also Gaussian and it is parameterized by its mean vector and a covariance matrix.

For {given} $\gamma, \omega$, the \emph{posterior} distribution 
\begin{equation}
\label{eq:posterior} p_{\gamma, \omega}(\phi \mid \D^{(i)}) = \frac{p_\gamma(\mathbf{y}^{(i)} \mid \mathbf{u}^{(i)}, \phi) p_\omega(\phi)}{p_{\gamma, \omega}(\mathbf{y}^{(i)} \mid \mathbf{u}^{(i)})} 
\end{equation}
describes the posterior belief of the latent parameters of system $(i)$, given the observation of $\D^{(i)} = (\mathbf{u}^{(i)}, \mathbf{y}^{(i)})$. 

However,~\eqref{eq:posterior} is a meaningful characterization of uncertainty only 
if the prior and likelihood accurately describe the generative process underlying 
the meta-dataset $\mathcal{D}$. The meta-parameters $\gamma, \omega$ must 
therefore be optimized accordingly.
Moreover, because the base architecture $F$ is a neural network, the marginal likelihood, or \emph{evidence} $p_{\gamma, \omega}(\mathbf{y}^{(i)} \mid \mathbf{u}^{(i)}) = \int_\phi {p_\gamma(\mathbf{y}^{(i)} \mid \mathbf{u}^{(i)}, \phi) p_\omega(\phi)}{p_{\gamma, \omega}} \; d\phi$ 
appearing in the denominator of~\eqref{eq:posterior} is intractable, and consequently 
the posterior cannot be computed analytically.  To address both challenges, we rely on Variational Inference~\cite{blei2017variational}.
\begin{remark}
The Gaussian prior assumption is not restrictive, as the nonlinear transformations 
induced by $F$ and $P_\gamma$ can map this simple prior into arbitrarily complex 
distributions over the output space. Moreover, one can always choose a \emph{fixed} 
isotropic Gaussian prior $\mathcal{N}(\mathbf{0}, \mathbf{I})$, since the effect of 
a different mean and covariance can be absorbed into the first layer of $P_\gamma$. 
We adopt this isotropic prior in the remainder of the paper, and accordingly 
omit the prior parameters $\omega$.
\end{remark}

\subsubsection{Amortized Variational Inference and the ELBO}
The core idea of Variational Inference is to approximate the intractable posterior $p_\gamma(\phi \mid \D^{(i)})$ with a parameterized \emph{variational} distribution $q_\psi$.  In this paper, $q_\psi$ is chosen as a multivariate Gaussian with diagonal covariance. 
To avoid learning a separate set of variational parameters for each dataset in our meta-dataset, we employ \textit{amortized} variational inference. We introduce an inference network (encoder) $E_\psi$, with weights $\psi$, that takes as input a dataset and outputs the sufficient statistics of the approximate posterior. Specifically, taking into account the train-test split within each dataset $\D^{(i)} = (\D_{\tr}^{(i)}, \D_{\te}^{(i)})$, the encoder processes the training portion $\D_{\tr}$ and provides a mean vector and the diagonal elements of the covariance: $E_\psi(\D_{\tr}^{(i)}) = (\boldsymbol{\mu}^{(i)}, \boldsymbol{\sigma}^{(i)})$. The variational distribution $q_\psi(\phi \mid D^{(i)}_\tr)$ is $\mathcal{N}(\boldsymbol{\mu}^{(i)}, \diag(\boldsymbol{\sigma}^{(i)}))$.

In Variational Inference, the Kullback-Leibler (KL) divergence between the approximate and the exact posterior is minimized. Mathematically, this is equivalent to maximizing (w.r.t. the variational parameters $\psi$) the Evidence Lower Bound (ELBO)~\cite{doersch2016tutorial}, that for a single dataset is defined by:
\begin{multline}
\label{eq:classic_elbo}
\text{ELBO}^{(i)}(\gamma, \psi) = 
\mathbb{E}_{q_{\psi}(\phi \mid \D_{\tr}^{(i)})}\left[ \log p_\gamma(\mathbf{y}_\te^{(i)} \mid \mathbf{u}_\te^{(i)}, \phi) \right] \\
- \text{D}_{\text{KL}}\left(q_{\psi}(\phi \mid \D_{\tr}^{(i)}) \parallel \mathcal{N}(\mathbf{0}, \mathbf{I})\right),
\end{multline}

Crucially, the ELBO is a lower bound of the \emph{log-evidence} $\log p_{\gamma}(\mathbf{y}^{(i)} \mid \mathbf{u}^{(i)})$, and the gap is the KL divergence between $q_{\psi}(\phi \mid \D_{\tr}^{(i)})$ and $p_{\gamma}(\phi \mid \D_{\tr}^{(i)})$, namely: 
\begin{multline}
\log p_{\gamma}(\mathbf{y}^{(i)} \mid \mathbf{u}^{(i)}) = \text{ELBO}^{(i)}(\gamma, \psi) 
+\\
\text{D}_{\text{KL}}\left(q_{\psi}(\phi \mid \D_{\tr}^{(i)}) \parallel p_{\gamma}(\phi \mid \D_{\tr}^{(i)})
\right)
\end{multline}
Therefore, by \emph{jointly} maximizing the ELBO with respect the variational parameters $\psi$ and the generative model parameters $\gamma$, it is possible to recover (i) a probabilistic model that maximizes the data evidence, and thus is optimal in the (marginal) maximum likelihood sense and (ii) as a by-product, an approximate expression of the otherwise intractable posterior~\eqref{eq:posterior}. 

\begin{remark}
The ELBO expression \eqref{eq:classic_elbo} balances two competing objectives. The first term is the expected log-likelihood (reconstruction error), which encourages the latent representation 
to accurately reproduce the output dynamics via the lifting function. The second term acts as a regularizer, preventing the approximate posterior $q_{\psi}(\phi \mid \D^{(i)})$ from deviating too much from the prior.
\end{remark}

\subsubsection{Meta-training}
By maximizing the ELBO across the dataset distribution $p(\D)$, we learn a generative model of $p(\D)$ governed by a low number of latent factors $\phi$, aligning with the modeling assumption~\eqref{eq:data_distrib}. Overall, the \textit{meta-ELBO} objective is:
\begin{multline}
\label{eq:meta_elbo}
\mathcal{J}(\gamma, \psi) = \mathbb{E}_{p(\D)} [ \mathbb{E}_{q_{\psi}(\phi \mid \D_{tr})} \left[ \log p_\gamma(\mathbf{y}_{te} \mid \mathbf{u}_{te}, \phi) \right] \\ 
- \beta \text{D}_{\text{KL}} \left(q_{\psi}(\phi \mid \D_{tr}) \parallel \mathcal{N}(\mathbf{0}, \mathbf{I}) \right).
\end{multline}
Here, the hyperparameter $\beta$ (standard in $\beta$-VAEs~\cite{higgins2017betavae}) scales the relative weight of the KL-divergence regularization. Tuning $\beta$ proved critical in our experiments to actively prevent posterior collapse~\cite{lucas2019understanding} and ensure a meaningful latent representation. During training, the outer expectation over the dataset distribution is approximated via Monte Carlo sampling across mini-batches of size $b=128$. For the inner expectation, we draw a single stochastic sample per dataset from the approximate posterior $q_\psi$. To ensure that gradients can back-propagate through this random sampling step into the encoder parameters $\psi$, we employ the standard reparameterization trick~\cite{kingma2013auto}. This enables end-to-end gradient-based optimization of the joint meta-parameters:
\begin{equation}
\label{eq:meta_training}
\hat{\gamma}, \hat{\psi} = \arg\max_{\gamma, \psi} \mathcal{J}(\gamma, \psi).
\end{equation}

\begin{figure*}[htbp]
    \centering
    \includegraphics[width=\linewidth]{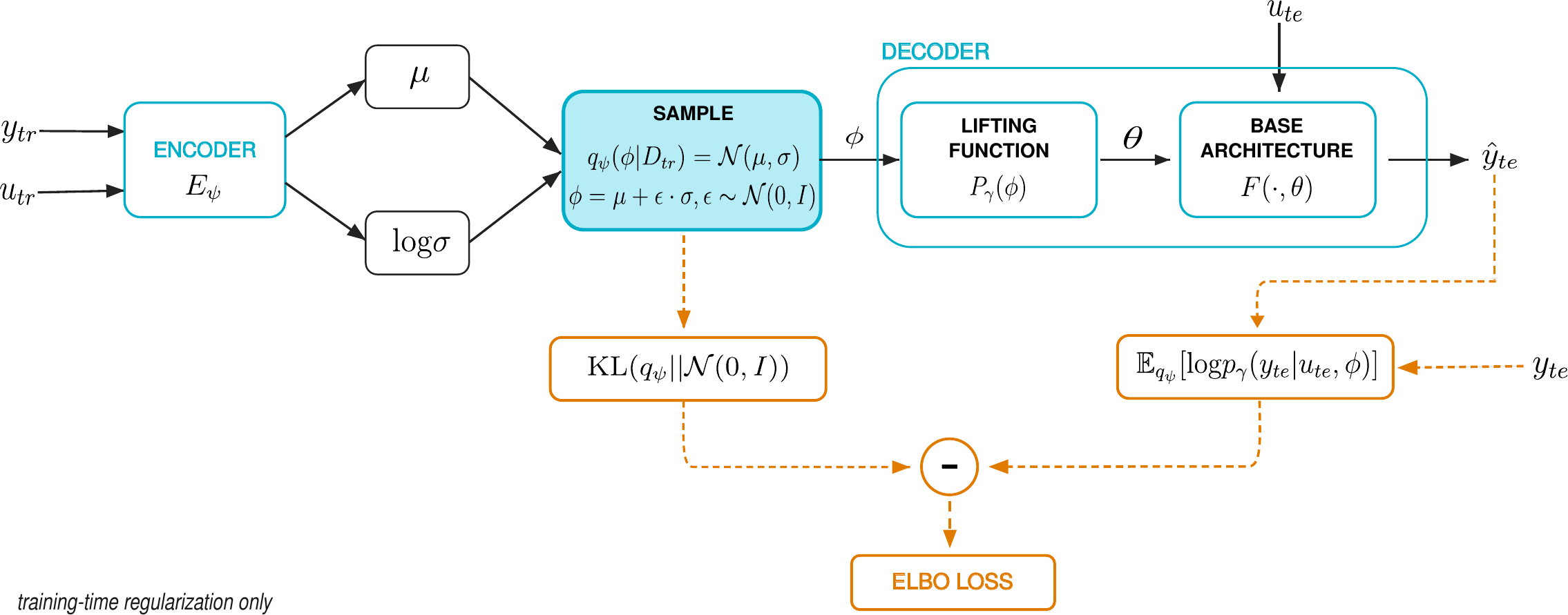}
    \caption{Overview of the meta-training pipeline. The encoder $E_\psi$ maps a training dataset to ($\mu$, log $\sigma$); $\phi$ is sampled via reparameterization and passed through the lifting function $P_\gamma$ and base architecture $F$ to produce $\hat{y}_{te}$. The KL term (orange, left) regularizes only at training time; the reconstruction term (orange, right) uses $y_{te}$.}
    \label{fig:pipeline}
\end{figure*}

The full meta-training framework pipeline is illustrated in Figure~\ref{fig:pipeline}.

\subsection{Meta-Testing, MAP and Laplace Approximation}
\label{sec:MAP}
At meta-test time, the framework is evaluated on a new, unseen system generating a dataset $\D^{\rm new} = \{\D^{\rm new}_\tr, \D^{\rm new}_\te\}$. While the encoder $E_{\hat{\psi}}$ provides an immediate, zero-shot Gaussian approximation of the posterior $q_{\hat \psi}(\phi \mid \D^{\rm new}_\tr)$, relying solely on this amortized network can lead to sub-optimal accuracy and miscalibrated uncertainty for out-of-distribution dynamics. 
To refine the prediction, we first perform Maximum A Posteriori (MAP) estimation, using the encoder's mean output as an initialization point. 
Specifically, we obtain the MAP point estimate $\phi_{\text{MAP}}$ as:

\begin{subequations}   
\label{eq:map_lap}
\begin{equation}
\label{eq:map}
\hat \phi = \arg \min_\phi \mathcal{L^{\rm nlp}}(\phi),
\end{equation}
where the loss $\mathcal{L^{\rm nlp}(\phi)} = \norm{\frac{\mathbf{y}_\tr^{\rm new} - F(\mathbf{u}^{\rm new}_\tr; P_{\hat{\gamma}}(\phi))}{\sigma_e}}^2 + \norm{\phi}_2^2$ is the negative log-posterior, up to an additive factor. 
The optimization mirrors the task-specific fine-tuning of the deterministic framework~\cite{forgione2025manifold}, yielding a single point estimate for the low-dimensional parameter.

Then, to recover an accurate uncertainty measure, we apply the Laplace approximation around the MAP estimate. By taking the second-order Taylor expansion of the log-posterior at $\phi_{\text{MAP}}$, we approximate the posterior as a Gaussian $\mathcal{N}(\phi_{\text{MAP}}, \Sigma_{\text{Laplace}})$, where the covariance matrix is the inverse of the Hessian of the negative log-posterior loss: 
\begin{equation}
\Sigma_{\text{Laplace}} = \left( \left . \frac{\partial^2 \mathcal{L^{\rm nlp}(\phi)}}{\partial \phi^2} \right |_{\phi = \phi_{\text{MAP}}} \right)^{-1}.
\label{eq:uncertainty_est}
\end{equation}
\end{subequations}

Because the dimensionality of the manifold $n_\phi$ is small, computing and inverting this Hessian is computationally feasible. This yields a principled, fast, and mathematically grounded quantification of uncertainty.

\begin{remark}
The Laplace approximation in~\eqref{eq:uncertainty_est} is a \emph{local}
approximation: it fits a single Gaussian around the mode
$\phi_{\mathrm{MAP}}$ found by~\eqref{eq:map}, and by construction
cannot represent multimodality or heavy-tailed structure in the true
posterior $p_\gamma(\phi \mid \mathcal{D})$. This is a meaningful
restriction whenever distinct latent configurations $\phi$ can explain
the same observed trajectory equally well, a situation that is not
merely theoretical for hysteretic systems such as Bouc--Wen, where several
combinations of the shape parameters $(\gamma,\beta,\delta,\nu)$ can
produce near-identical force--deformation loops, particularly under short
or narrowband excitation. In such regimes the single-mode Gaussian
in~\eqref{eq:uncertainty_est} may underestimate total predictive
uncertainty by ignoring probability mass around competing modes, or
misrepresent the local curvature if $\phi_{\mathrm{MAP}}$ lands near a
saddle between two comparably good optima. The empirical coverage results
in Fig.~\ref{fig:sub_b}, evaluated on the nominal-parameter benchmark
trajectory, do not stress-test this failure mode directly, since the
adaptation data there is long enough ($L \geq 100$) to identify a
well-separated mode; shorter or less-informative excitations are where
multimodal collapse is most likely. Addressing this is a natural direction
for future work, e.g., replacing the single Laplace posterior with a
mixture of Laplace approximations fit at multiple MAP restarts, in the
spirit of~\cite{eschenhagen2021mixtures}, or a full-covariance variational posterior that is not restricted to a
diagonal Gaussian at inference time.
\end{remark}

\section{Numerical Examples}
\label{sec:results}

The proposed variational framework is evaluated on two examples: a synthetic toy example concerning static regression of sinusoidal functions, and a dynamical system identification problem using the Bouc--Wen benchmark \cite{noel2016hysteretic}. In both cases, the variational approach is compared against its deterministic counterpart \cite{forgione2025manifold}. The primary objective is to demonstrate that the probabilistic formulation achieves comparable predictive performance while additionally providing mathematically rigorous, well-calibrated uncertainty estimates for both the output predictions and the low-dimensional latent parameters.

The code was implemented in JAX and is publicly available in the Github repository \url{https://github.com/mattrufolo/VAE-sysid-neural-manifold}. All computations were executed on a server equipped with an NVIDIA RTX 3090 GPU. The network structures used in the numerical examples are summarized in Table~\ref{tab:architectures}.

\begin{table*}[t]
    \centering
    \caption{Architectures employed in the numerical examples.}
    \label{tab:architectures}
    \renewcommand{\arraystretch}{1.3} 
    \begin{tabular}{@{}lp{4.5cm}p{3.5cm}p{4.5cm}@{}}
        \toprule
        \textbf{Benchmark} & \textbf{Encoder ($E_\psi$)} & \textbf{Lifting Function ($P_\gamma$)} & \textbf{Base Architecture ($F$)} \\
        \midrule
        \textbf{Sines} & 
        \textbf{DeepSets:} \newline
        $\bullet$ \textit{Shared MLP:} 2 hidden layers, ReLU \newline
        $\bullet$ \textit{Aggregation:} Mean pooling \newline
        $\bullet$ \textit{Output MLP:} 1 hidden layer, ReLU $\to$ parallel Dense layers for $\boldsymbol{\mu}$ and $\log\boldsymbol{\sigma}$ & 
        \textbf{Affine Map:} \newline
        Dense layer (no activation) & 
        \textbf{MLP:} \newline
        Hidden dimensions: [64, 32, 16, 1], ReLU activations \\
        \midrule
        \textbf{Bouc-Wen} & 
        \textbf{RNN + MLP:} \newline
        $\bullet$ \textit{RNN:} Bi-directional GRU (128 units) \newline
        $\bullet$ \textit{Aggregation:} Average pooling \newline
        $\bullet$ \textit{Output MLP:} 1 hidden layer (128 units), tanh $\to$ parallel Dense layers for $\boldsymbol{\mu}$ and $\log\boldsymbol{\sigma}$ ($n_\phi=20$) & 
        \textbf{Affine Map:} \newline
        Dense layer (no activation), expanding $n_\phi=20$ to $n_\theta=244$ & 
        \textbf{Neural State-Space:} \newline
        $\bullet$ $n_x=3, n_u=1, n_y=1$ \newline
        $\bullet$ $N_f, N_g$: MLPs with 1 hidden layer (16 units), tanh activations \\
        \bottomrule
    \end{tabular}
\end{table*}

\subsection{Static regression}
The meta-training dataset comprises $1000$ sine wave regression tasks of length $N = 100$. For each dataset $\D^{(i)}$, outputs are generated from the sinusoidal function over the input with an additive white noise, via $\mathbf{y}^{(i)} = A^{(i)} \sin(\mathbf{u}^{(i)} + \alpha^{(i)}) + \mathbf{e}^{(i)}$, with inputs $\mathbf{u}^{(i)} \sim \mathcal{U}(-5, 5)$, amplitudes $A^{(i)} \sim \mathcal{U}(0.5, 2.0)$, and phases $\alpha^{(i)} \sim \mathcal{U}(0, \pi)$. The term $e^{(i)}$ denotes additive white noise with $1\%$ relative amplitude. Each sequence is evenly split into training and test sets $\D^{(i)} = \{\D_{\tr}^{(i)}, \D_{\te}^{(i)}\}$ ($N_\tr = N_\te = 50$). Because the underlying generative process has exactly two degrees of freedom (amplitude and phase), the latent manifold dimensionality is fixed to the theoretical minimum $n_\phi = 2$.

Architectural details are provided in Table~\ref{tab:architectures}. We employ a permutation-invariant DeepSets encoder \cite{zaheer2017deep}, aligning with the structural prior of static regression. 
Meta-training is performed by minimizing~\eqref{e1:amortized_train} and~\eqref{eq:meta_training} for the deterministic and probabilistic settings, respectively, using the Adam optimizer with a learning rate of $10^{-2}$ over $10{,}000$ iterations. To ensure a fair comparison, all hyperparameters and architectural choices are shared between the two frameworks, with the sole addition of the ELBO regularization weight $\beta=0.1$ for the probabilistic setting.

\subsubsection{Unconditional Generation from the Prior}
\begin{figure}[htbp]
    \centering
    \includegraphics[width=0.7\linewidth]{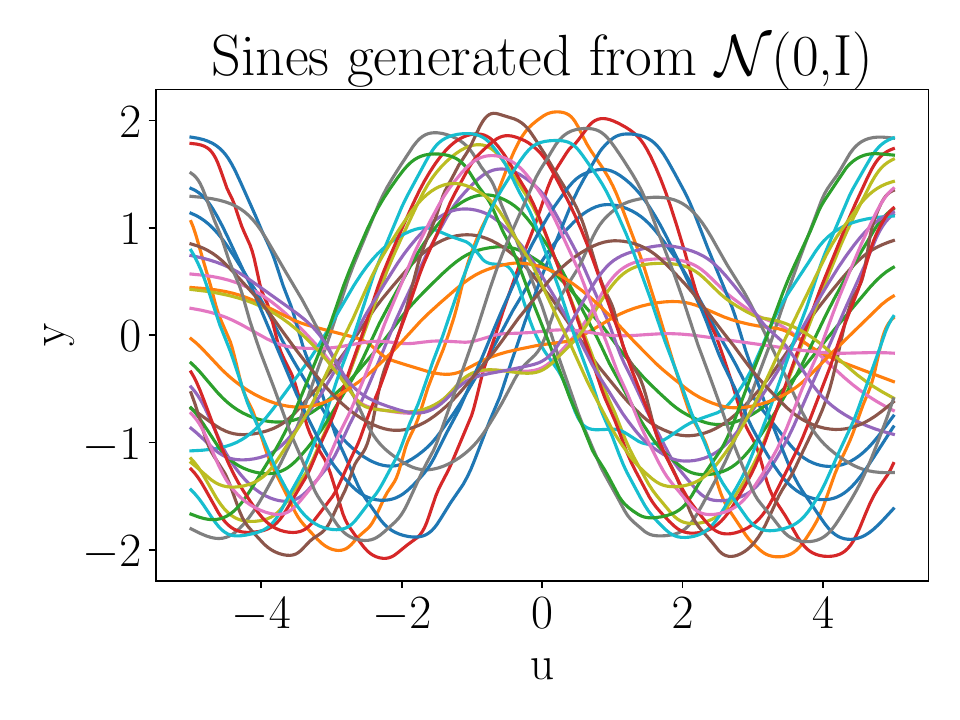}
    \caption{Unconditional generation of $30$ trajectories from the prior $\phi \sim \mathcal{N}(\mathbf{0}, \mathbf{I})$. The outputs exhibit the expected sinusoidal pattern characterizing the meta-training distribution.}
    \label{fig:unconditional_sines}
\end{figure}

We first assess the ability of the learned probabilistic model to generate synthetic datasets that resemble the meta-dataset. To this aim, we draw $30$ latent vectors samples $\tilde{\phi}^{(i)}$ from the prior $\mathcal{N}(\mathbf{0}, \mathbf{I})$ and evaluate the functions $F(\cdot, P_{\hat \gamma}(\tilde{\phi}^{(i)}))$ over a dense grid $\mathbf{u}_\mathrm{ds}$ of $1000$ linearly spaced points in the domain $[-5, 5]$.
As shown in Figure~\ref{fig:unconditional_sines}, these unconditionally-sampled parameters generate plausible trajectories that exhibit a clear sinusoidal pattern. For a quantitative verification, we fit a parametric sinusoidal model to these generated curves to extract their empirical amplitudes and phases. We obtained generated amplitudes $\hat{A} \in [0.332, 2.101]$ and phases $\hat{\varphi} \in [0.006, 3.165]$. These empirical ranges are aligned with the ranges 
$[0.5, 2.0]$ and $[0, \pi]$ of the uniform distributions of $A$ and $\alpha$ used to generate the meta-training set. This illustrates that the learned prior effectively captures the true generative process of the sines meta-dataset.

\subsubsection{Performance and Uncertainty in Testing}
During meta-testing, the shared meta-learned representation is used to find the optimal low-dimensional parameters $\phi$ that fit the training portion of an unseen dataset. This adaptation is implemented for the deterministic and probabilistic cases by solving Equations~\eqref{eq:inner} and \eqref{eq:map_lap}, respectively. This procedure was executed across 100 new test trajectories. To evaluate the robustness of the adaptation, we assessed the models under three different limited-data regimes by varying the length of the training sequence ($N_\tr=1$, $N_\tr=2$, and $N_\tr=5$). To measure the true functional accuracy of the adapted models, the predictive Root Mean Square Error (RMSE) is computed over the same high-resolution grid $\mathbf{u}_{ds}$ utilized to generate Fig.~\ref{fig:unconditional_sines}. The quantitative results, presented in Table~\ref{tab:results_sines} as the mean and standard deviation across all $100$ test trajectories, demonstrate that the predictive performance of the proposed VAE formulation remains highly competitive with the baseline deterministic approach.

\begin{table}[!ht]
\centering
\caption{Continuous functional RMSE for the Sines benchmark evaluated over a dense 1000-point grid. Results are reported as the mean $\pm$ standard deviation across 100 unseen test trajectories for different training sequence lengths ($N$).}
\label{tab:results_sines}
\small
\renewcommand{\arraystretch}{1.2} 
\begin{tabular}{@{}ccc@{}}
\toprule
\textbf{Training Length ($N$)} & \textbf{Deterministic RMSE} & \textbf{VAE RMSE} \\
\midrule
$N = 2$  & 0.548 $\pm$ 0.723 & 0.634 $\pm$ 0.909 \\
$N = 4$  & 0.121 $\pm$ 0.273 & 0.086 $\pm$ 0.180 \\
$N = 10$ & 0.062 $\pm$ 0.124 & 0.026 $\pm$ 0.044 \\
\bottomrule
\end{tabular} 
\end{table}
\begin{figure}[htbp]
    \centering
    \begin{minipage}{\linewidth}
        \centering
        \includegraphics[width=0.83\linewidth, keepaspectratio]{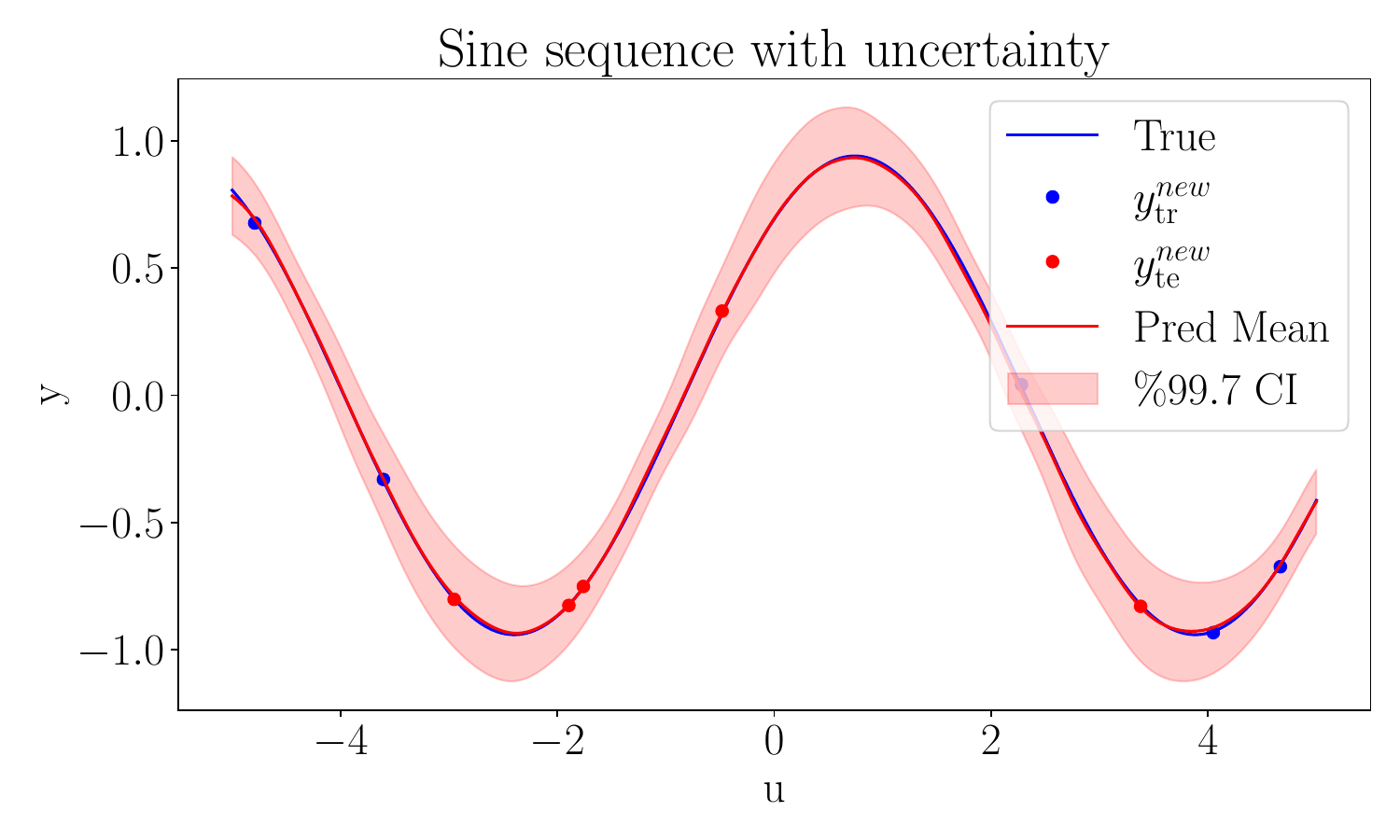}
    \end{minipage}
    
    
    \begin{minipage}{\linewidth}
        \centering
        \includegraphics[width=0.8\linewidth, keepaspectratio]{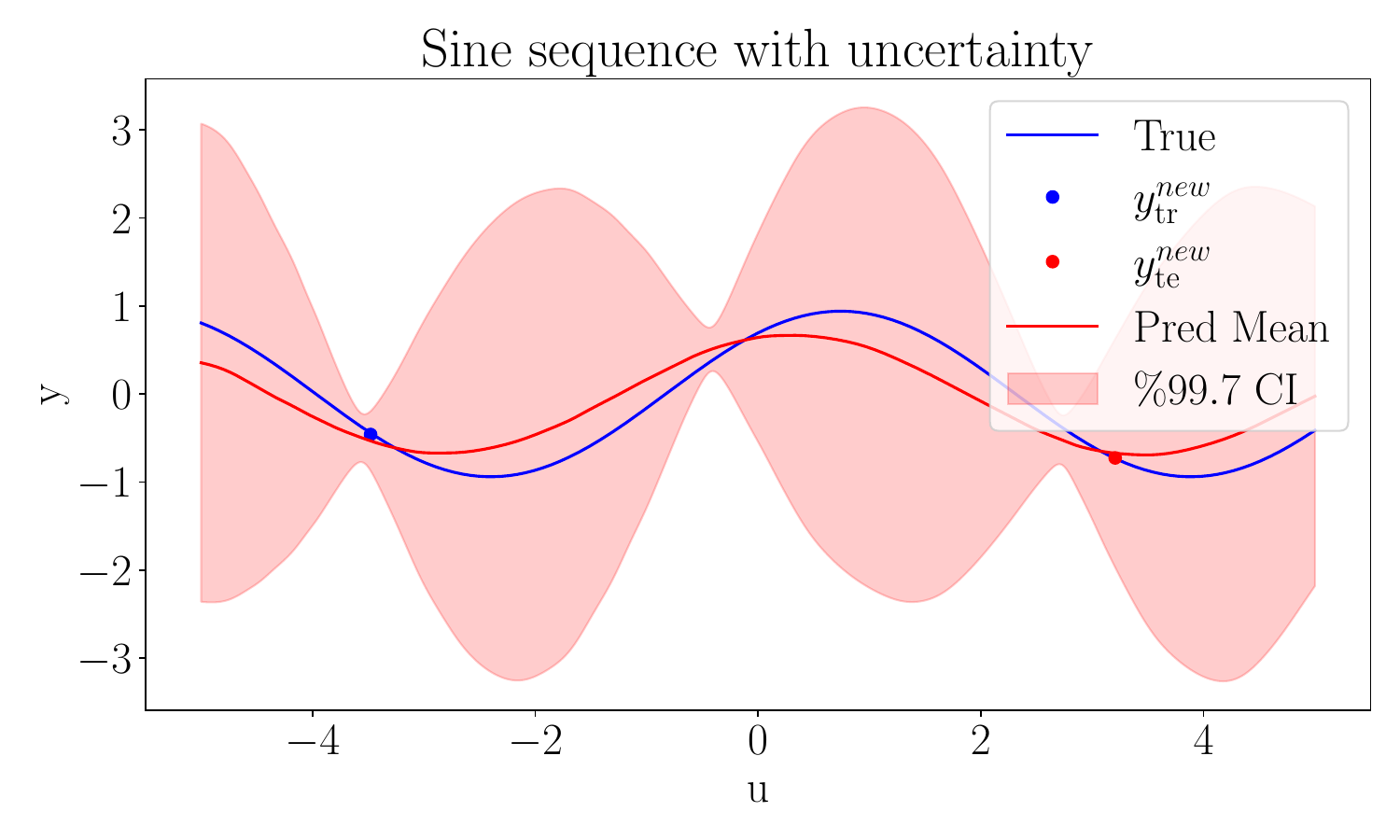}
    \end{minipage}
    
    \caption{Laplace uncertainty bounds on a novel test sequence. \textbf{Top:} Adaptation with 5 points ($N=10$) yields a tight fit. \textbf{Bottom:} Adaptation with 1 point ($N=2$) results in a poor pointwise fit, yet the $\pm 3\sigma$ bounds correctly maintain the inherent sinusoidal structure learned from the prior.}
    \label{fig:sines_uncertainty}
\end{figure}

The uncertainty quantification capability of the probabilistic meta-learning approach is visualized in Figure~\ref{fig:sines_uncertainty} for $N_\tr = 5$ and $N_\tr = 1$. In these plots, the discrete blue markers represent the sparse observations used for task adaptation, while the solid lines represent the continuous functional predictions evaluated over the high-resolution grid. For $N_\tr = 5$, the training points are sufficient to accurately anchor the phase and amplitude, yielding a tight fit and narrow uncertainty bounds. Conversely, for $N_\tr = 1$, adaptation relies on a single data point, leading to a predictably poor pointwise fit. However, the uncertainty bands in this data-starved regime remarkably retain a clear, bounded oscillatory structure. This confirms that the meta-learned probabilistic model successfully captured the fixed-frequency pattern across tasks, even when the data is insufficient to pinpoint the exact phase or amplitude.

\subsection{Bouc-Wen benchmark}
We perform a comparison between the probabilistic meta-learning framework introduced in this paper and its deterministic counterpart on the same Bouc-Wen benchmark~\cite{noel2016hysteretic} utilized in~\cite{forgione2025manifold}.
The benchmark models the vibrations of a 1-DoF hysteretic mechanical system whose dynamics are described in state-space form by:
\begin{subequations}  
\label{eq:boucwen}
\begin{align}
\begin{bmatrix} 
\dot{p}(t) \\ 
\dot{v}(t) \\ 
\dot{z}(t) 
\end{bmatrix}
&=
\begin{bmatrix} 
v(t) \\ 
\frac{1}{m_L} \left (u(t) -k_L p(t) - c_L v(t) - z(t) \right) \\ 
\alpha v(t) - \beta(\gamma |v(t)| |z(t)|^{\nu -1} + \delta v(t) |z(t)|^\nu)
\end{bmatrix}\\
y(t) &= p(t),
\end{align}
\end{subequations}
where $p(t)$~(m) is the measured position, $v(t)$~(m/s) the unmeasured velocity, $z(t)$~(N) an unmeasured hysteretic force, and $u(t)$~(N) the known input force. 

\begin{table}[!ht]
\setlength{\tabcolsep}{4pt}
\centering
\caption{Bouc--Wen coefficients: nominal and uniform min/max bounds.}
\label{tab:bw_params}
\small
\begin{tabular}{c||cccccccc}
\toprule
 & $m_L$ & $c_L$ & $k_L$ & $\alpha$ & $\beta$ & $\gamma$ & $\delta$ & $\nu$\\
\midrule
nom & 2 & 10 &$5.0\times 10^4$ & $5.0\times 10^4$ & 1000 & 0.8 & -1.1 & 1\\
min & 1 & 5 & $2.5\times 10^4$ & $2.5\times 10^4$ & $500$ & 0.5 & -1.5 &1\\
max & 3 & 15 & $7.5\times 10^4$ & $7.5\times 10^4$ & 4500 & 0.9 & -0.5 &1\\
\bottomrule
\end{tabular} 
\end{table}

The benchmark provides a noise-free test dataset generated by simulating~\eqref{eq:boucwen} with the nominal coefficients in Table~\ref{tab:bw_params}, excited by a random-phase multisine with length $N=8192$, sampled at frequency $f_s=750$~Hz, that excites the frequency range $[5, 150]$~Hz, and has as root mean square of $50$~N. A training dataset ($N=40960$) is also provided, where the system is excited by a distinct multisine signal with the same properties, and the output is corrupted by additive Gaussian noise (bandwidth $[0-375]$~Hz, with amplitude $8 \cdot 10^{-3}$). 

Following~\cite{chakrabarty2025meta}, we extend the original scope of the benchmark and use~\eqref{eq:boucwen} to generate a \emph{meta-training} dataset composed of random multisine trajectories from systems with coefficients uniformly sampled within the min-max bounds in Table~\ref{tab:bw_params}. 
Crucially, to prevent meta-overfitting, the reduced-complexity architecture is optimized exclusively on this synthetic meta-dataset. The original benchmark dataset (simulated at nominal values) is strictly reserved as \emph{meta-test} set to evaluate the final adaptation performance.

\subsection{Meta-Training the System Manifold}
\label{sec:meta_training}
To enable robust identification in the low-data regime, we meta-learn a low-dimensional manifold ($n_\phi=20$) embedded within the full-complexity parameter space ($n_\theta=244$). The specific configurations for the encoder $E_\psi$, lifting function $P_\gamma$, and base architecture $F$ are detailed in Table~\ref{tab:architectures}. 

The lifting network is affine: $P_\gamma(\phi) = V \phi + \theta_{\rm bias}$, with tunable parameters $\gamma = \mathrm{vec}(V, \theta_{\rm bias}) \in \mathbb{R}^{5124}$. The encoder $E_\psi$, which comprises $n_\psi = 136,340$ parameters, utilizes a Bi-Gated Recurrent Unit (GRU)~\cite{chung2014empirical} followed by an average-pooling mechanism to aggregate input-output sequences into a compact latent representation.
We use as base architecture the same linear-plus-residual neural state-space model from~\cite{forgione2025manifold}:
\begin{subequations}
\label{eq:ss_boucwen}
\begin{align}
    x_{k+1} &= A x_k + B u_x + N_f(x_k, u_k; W_f) \\
    y_{k} &= C x_k +  N_g(x_k; W_g),
\end{align}
\end{subequations}
with $n_\theta = 244$ parameters (further details in Table~\ref{tab:architectures}).

For the probabilistic setting, we use $\beta = 0.1$ in \eqref{eq:meta_elbo}.  The meta-dataset $\mathcal{D}$ consists of input-output sequences of length $N=2000$, generated on-the-fly by sampling Bouc--Wen systems within the bounds of Table~\ref{tab:bw_params} and exciting them with random multisines. Optimization is performed using \texttt{Adam}, running for $200,000$ iterations with a learning rate decayed from $1 \cdot 10^{-4}$ to $1 \cdot 10^{-5}$ via cosine scheduling. Because datasets are generated dynamically, the meta-training is performed over 25.6 million unique trajectories and took $\sim$ 25.3 hours. 

\subsection{Task-Specific Adaptation and Uncertainty Analysis}
\label{sec:adaptation_training}
The primary objective of this numerical evaluation is to compare the proposed variational framework against its deterministic counterpart \cite{forgione2025manifold}. For a broader analysis comparing the baseline manifold meta-learning strategy against other state-of-the-art system identification methods,
we refer the reader to~\cite{forgione2025manifold}. Here, we focus exclusively on assessing how the probabilistic extension affects predictive accuracy and enables uncertainty quantification. In line with~\cite{forgione2025manifold}, the performance is evaluated in terms of the $\mathrm{FIT}$ index.

To systematically evaluate the proposed variational framework in the low-data regime, we execute a comprehensive Monte Carlo adaptation study. Following the experimental setup established in \cite{chakrabarty2025meta}, we extract $n_{\rm mc}=100$ independent subsequences of varying lengths $L \in \{100, 200, 400, 500, 600, 800, 1000, 2000, 3000, 4000, 5000\}$ from the official 40960-sample Bouc--Wen training dataset. For each subsequence, we adapt the meta-learned model to the specific system instance by performing MAP estimation over the low-dimensional latent space to identify the task-specific parameter $\phi_{\text{MAP}}$. 

\begin{figure}[!tb]
    \centering
    \includegraphics[width=\linewidth]{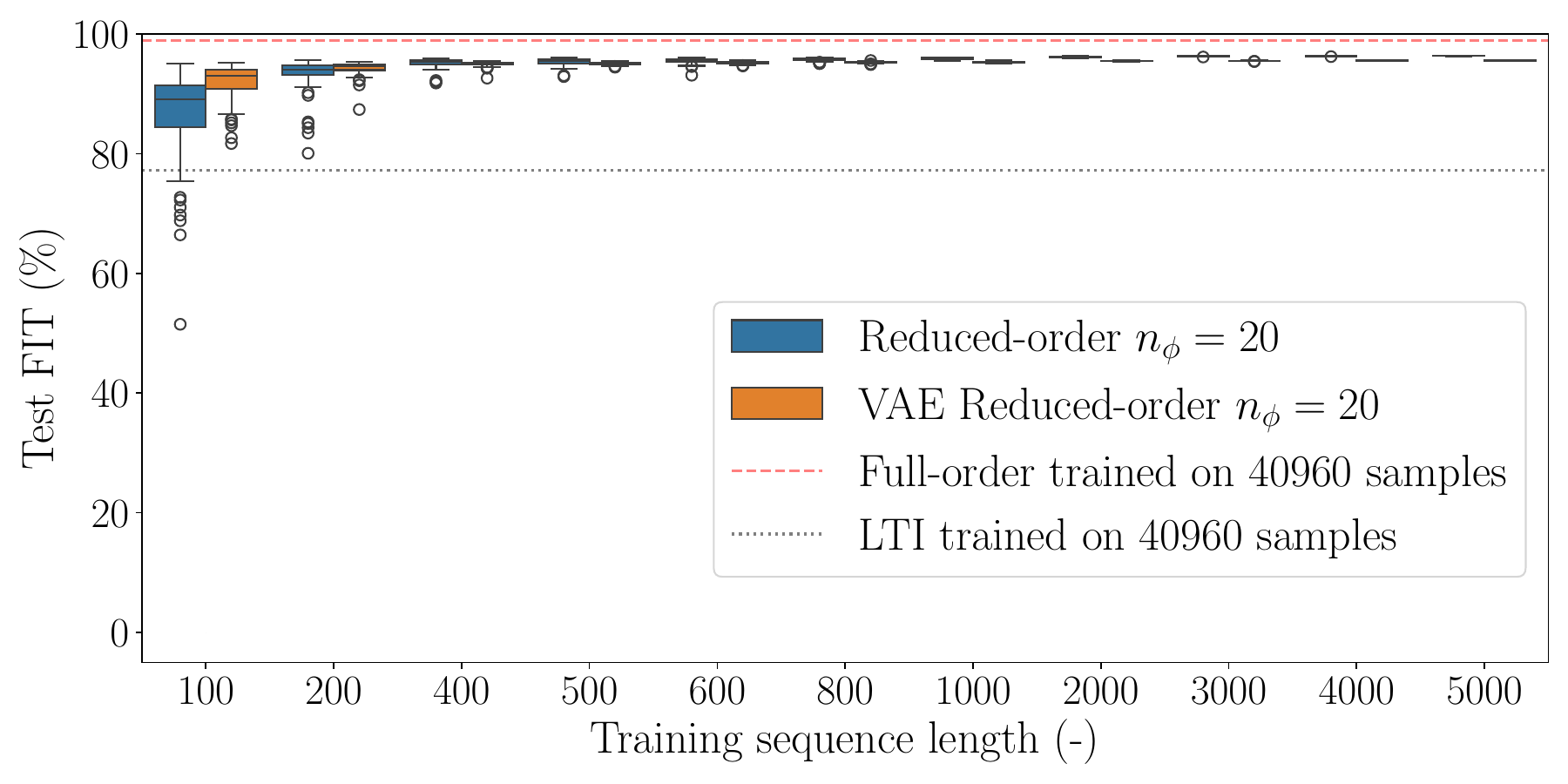}
    \caption{Deterministic and VAE reduced-order models: test FIT vs. training sequence length $L$.}
    \label{fig:boucwen_boxplot}
\end{figure}

Our evaluation serves two primary objectives. First, we demonstrate that transitioning to a probabilistic generative formulation does not sacrifice the predictive accuracy of purely deterministic methods. 
Figure~\ref{fig:boucwen_boxplot} reports the test $\mathrm{FIT}$ indices evaluated on the benchmark dataset. The results confirm that the variational framework remains highly competitive with its deterministic counterpart across all data regimes. To provide comprehensive context, we include as baselines an LTI and a full-complexity model~\eqref{eq:boucwen} trained on the complete 40960-sample training dataset; detailed formulations for these baselines are in~\cite{forgione2025manifold}.

\begin{figure}[htbp]
    \centering
    \begin{subfigure}{\linewidth}
        \centering
        \includegraphics[width=\linewidth, keepaspectratio]{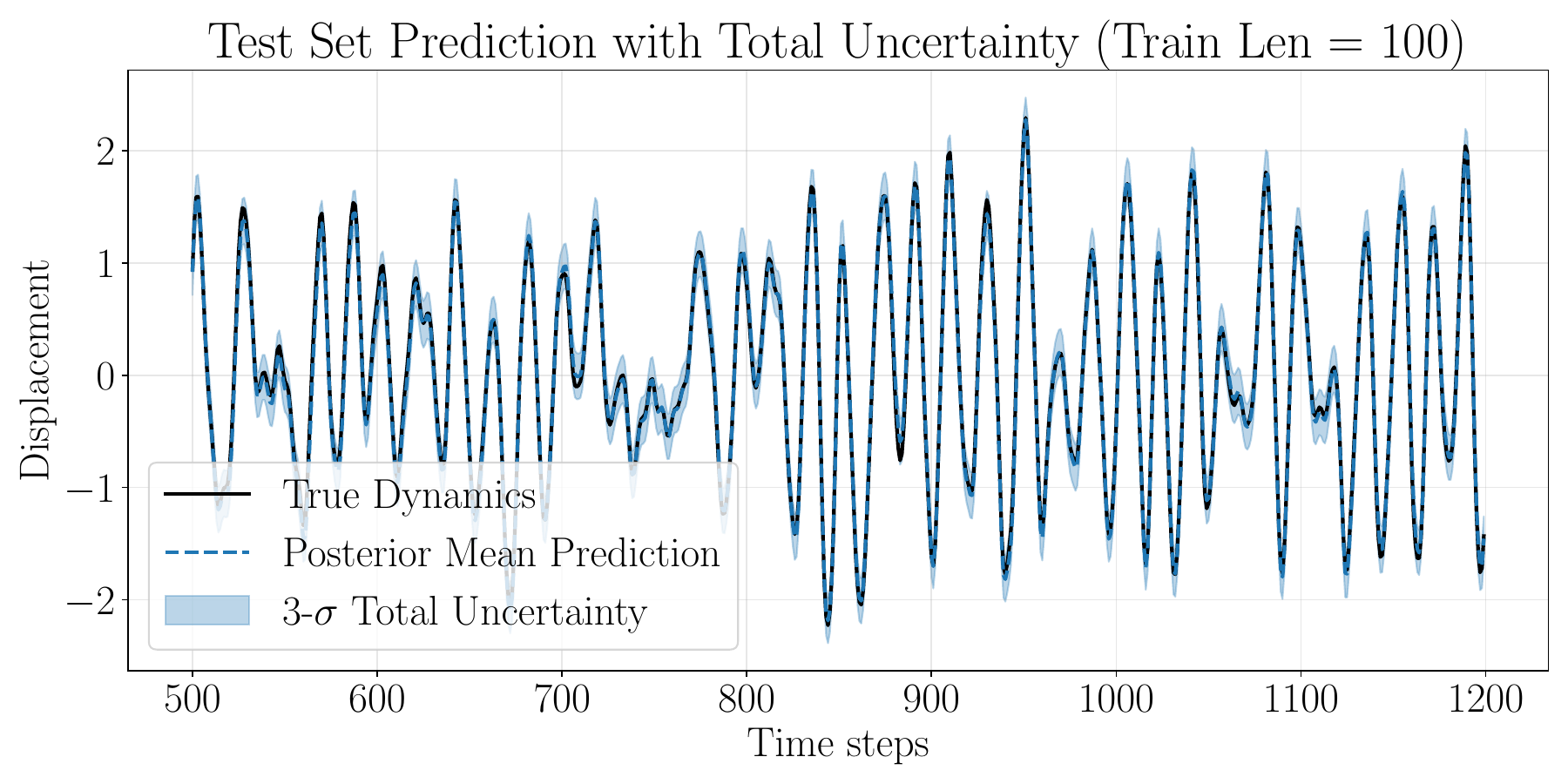} 
        \caption{Time-domain prediction with $\pm 3\sigma$ uncertainty bounds.}
        \label{fig:sub_a}
    \end{subfigure}
    
    \vspace{0.2cm} 
    
    \begin{subfigure}{\linewidth}
        \centering
        \includegraphics[width=\linewidth, height=3.5cm]{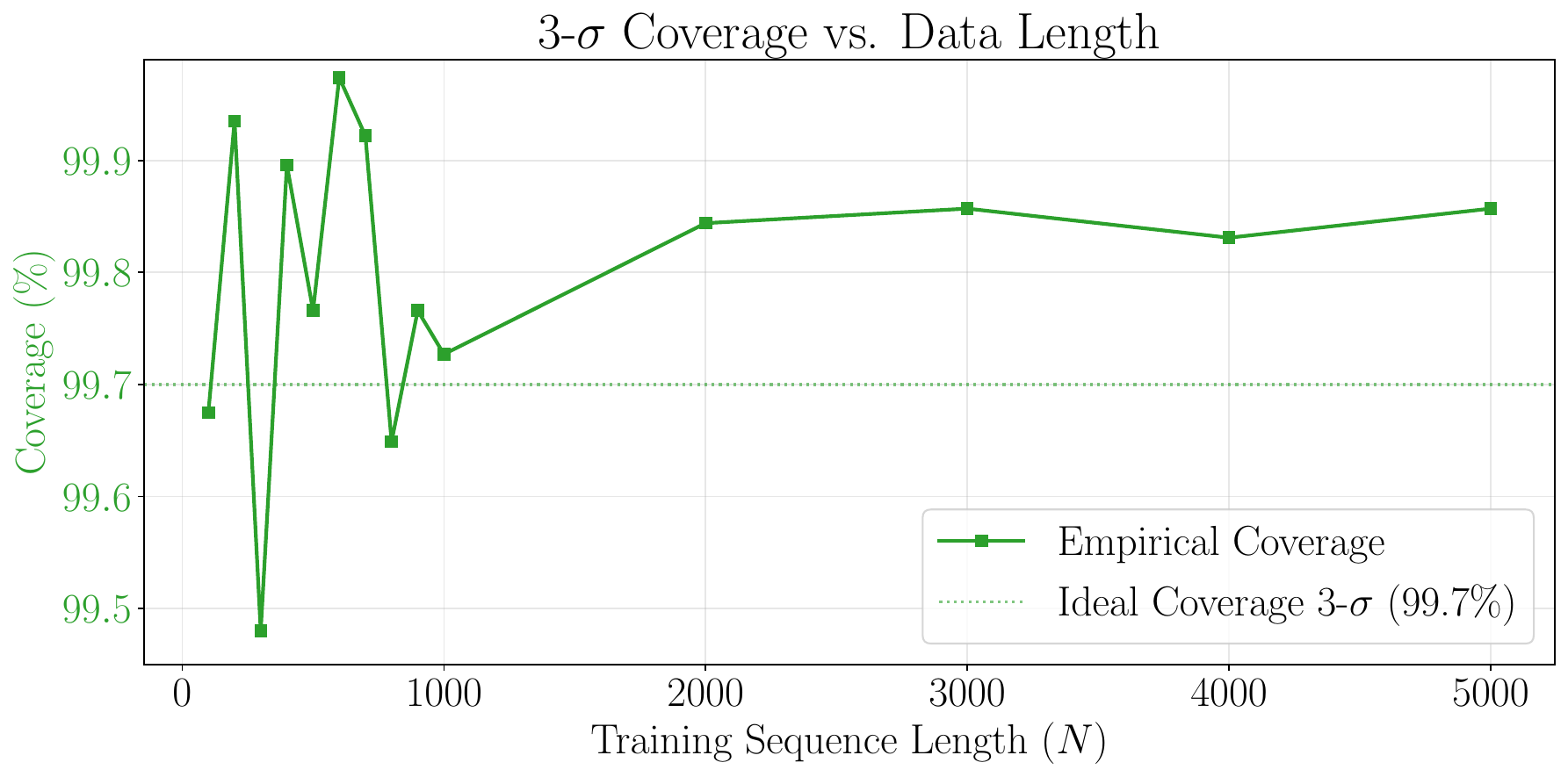} 
        \caption{Empirical coverage probability vs. adaptation length $L$.}
        \label{fig:sub_b}
    \end{subfigure}
    \caption{Performance and uncertainty validation of the variational framework. \textbf{(a)} Benchmark test prediction displaying calibrated confidence intervals after MAP adaptation and Laplace uncertainty computation ($L=100$). \textbf{(b)} Evaluation of empirical coverage against theoretical probability across different adaptation regimes.}
    \label{fig:boucwen_pred_uncertainty}
\end{figure}

\begin{figure}[htbp]
    \centering
    \includegraphics[width=\linewidth]{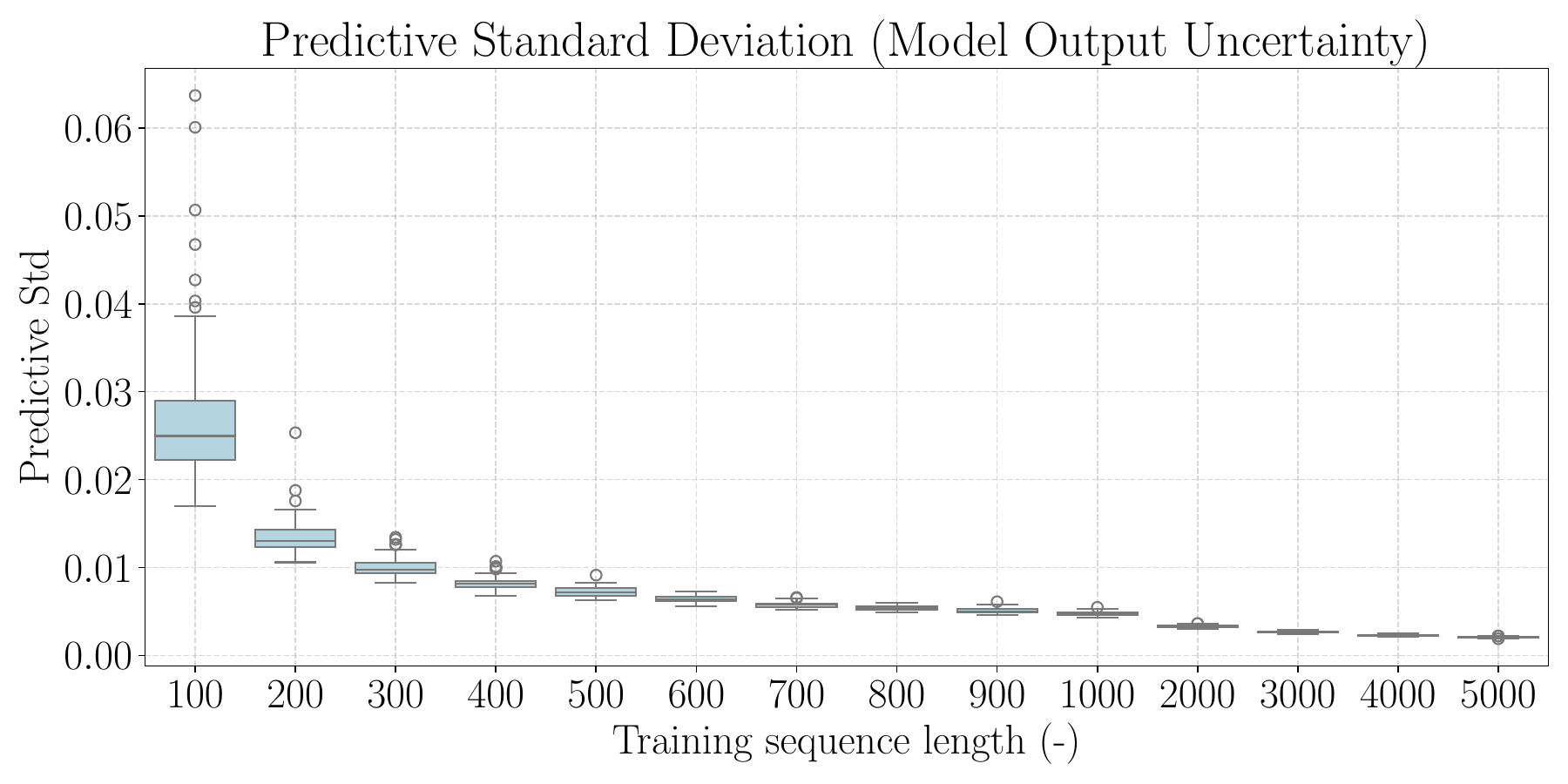}
    \caption{Evolution of model output uncertainty as a function of the adaptation sequence length.}
    \label{fig:std_decay}
\end{figure}

Second, and most critically, the variational approach equips the identified models with a mathematically rigorous quantification of epistemic uncertainty. By computing the inverse Hessian of the adaptation loss at $\phi_{\text{MAP}}$ via the Laplace approximation, we extract a localized Gaussian posterior for the latent system parameters. To illustrate the practical utility of this formulation, Figure~\ref{fig:sub_a} displays a benchmark test prediction alongside its $\pm 3\sigma$ confidence intervals, obtained after adapting the meta-model via MAP estimation with $L = 100$. The generated bounds successfully encapsulate the true system dynamics, actively compensating for regions where the mean prediction slightly deviates from the ground truth. Furthermore, to explicitly validate the statistical consistency of these bounds, Figure~\ref{fig:sub_b} evaluates the empirical coverage against the expected theoretical probability across various adaptation lengths, confirming that the framework yields properly calibrated uncertainty estimates.
Finally, to explicitly analyze how this uncertainty scales with data availability, Figure~\ref{fig:std_decay} plots the sequence-averaged predictive standard deviation as a function of the adaptation length $L$. The mean standard deviation exhibits a clear, monotonic decay as $L$ increases. This behavior perfectly aligns with Bayesian principles: as the model is provided with more task-specific information, the epistemic uncertainty inherently shrinks, yielding progressively more confident and reliable parameter estimates. 
\section{Conclusion}
\label{sec:conclusions}

This paper introduced a fully probabilistic framework  of  manifold meta-learning  for nonlinear system identification. By formulating the amortized optimization procedure via Variational Inference, the architecture learns a generative prior over a low-dimensional system manifold. While maintaining the high predictive accuracy of its deterministic counterpart, the proposed variational approach  provides  calibrated epistemic uncertainty bounds over both the latent physical parameters and the predicted trajectories. 

The framework was systematically validated on an interpretable synthetic regression task and the Bouc-Wen dynamical benchmark. In both cases, the model demonstrated a consistent, monotonic reduction in predictive variance as adaptation data increased, ensuring robust and reliable identification even in severely low-data regimes. Future work will focus on integrating these data-driven uncertainty bounds into risk-aware control applications, such as stochastic Model Predictive Control, and embedding physics-informed structural priors into the generative latent space to further improve interpretability and performance.



\bibliographystyle{IEEEtran} 
\bibliography{biblio}


\end{document}